\RequirePackage{silence}
\documentclass[]{ceurart}

\usepackage{booktabs}
\usepackage{amsmath}
\usepackage{tikz}
\usetikzlibrary{arrows.meta,positioning,shapes.geometric}
\usepackage{listings}
\usepackage{placeins}
\newcommand{\breakableensemblepositionweighted}{\texttt{ensemble\_\allowbreak{}position\_\allowbreak{}weighted}}
\newcommand{\breakableensembletoptwounion}{\texttt{ensemble\_\allowbreak{}top2\_\allowbreak{}union}}
\newcommand{\breakableensembleweightedunion}{\texttt{ensemble\_\allowbreak{}weighted\_\allowbreak{}union}}

\usepackage{pifont} 
\newcommand{\mycheck}{\ding{51}} 
\newcommand{\mycross}{\ding{55}} 

\DeclareMathOperator*{\argmax}{arg\,max}

\ExplSyntaxOn
\RenewDocumentCommand \printemails { } {
  \group_begin:
  \int_compare:nTF { \g_ead_int > 0 } {
    \tex_let:D \thefootnote \relax \footnotetext {
      \raggedright
      \hspace*{-\parindent}
      \hspace*{-\footnotemargin}
      \bool_if:NTF \g_ceur_nologo_bool
        { \textsc{email:\c_space_token} }
        { \faIcon[regular]{envelope-open}\c_space_token }
      \seq_use:Nn \g_ceur_ead_seq { ;~ }
    }
  } { }
  \group_end:
}
\ExplSyntaxOff

\begin{document}

\copyrightyear{2026}
\copyrightclause{Copyright for this paper by its authors.
  Use permitted under Creative Commons License Attribution 4.0
  International (CC BY 4.0).}

\conference{CLEF 2026 Working Notes, 21--24 September 2026, Jena, Germany}

\title{FAU at ImageCLEF 2026 Task on Multimodal Reasoning: Robust Candidate Scoring and Concise Multilingual Visual Answering}

\author[1]{Mohamed Basem}[%
email={mohamed.elmashtooly@fau.de},
]
\author[1]{Vincent Christlein}[%
email={vincent.christlein@fau.de},
]
\address[1]{Friedrich-Alexander-Universit\"at Erlangen-N\"urnberg, Erlangen, Germany}

\begin{abstract}
We present our ImageCLEF 2026 Multimodal Reasoning system for the Visual Multiple Choice Question Answering (Visual MCQ) and Visual Open Question Answering (Visual OpenQA) subtasks. The challenge requires reliable reasoning over multilingual educational and scientific images with dense text, diagrams, charts, tables, formulas, and units, while enforcing strict answer formats. Our central finding is that robust output control is as important as model choice. For Visual MCQ, we replace fragile free-form generation with direct candidate label scoring from vision-language model logits, then combine complementary runs through score fusion and voting. For Visual OpenQA, we use image enhancement, concise final answer prompting, deterministic decoding, and targeted post-processing to remove reasoning traces and formatting artifacts. Without task-specific model training, our official submissions achieved third place in Visual MCQ with 0.7108 accuracy and first place in Visual OpenQA with 0.6488 COMET, 0.1391 BLEU, 0.2762 ROUGE L, and 0.2383 METEOR. The results highlight the practical value of inference engineering: careful scoring, ensembling, prompting, and cleanup can turn strong VLMs into reliable competition systems.
\end{abstract}


{\hfuzz=408pt\maketitle}

\section{Introduction}

Multimodal exam questions are difficult for vision-language models (VLMs)
because they combine several skills in one input: reading dense text, following
diagrams and tables, understanding formulas and units, and answering in a very
specific format. In practice, many errors are not pure reasoning failures. A
model may identify the right answer but lose credit by producing an explanation,
copying extra text, or answering in the wrong language.

The ImageCLEF 2026 Multimodal Reasoning task studies this problem through two
subtasks~\cite{ImageCLEF2026,%
ImageCLEFMultimodalReasoningTaskOverview2026}. Visual MCQ asks systems to
choose one option label from A--E across six languages, while Visual OpenQA asks
for a short free-form answer in the question language. These two settings expose
different weaknesses: MCQ rewards clean label selection, whereas OpenQA rewards
concise and semantically accurate generation.

Our main finding is simple: \emph{controlling the output matters as much as
choosing the model}. For MCQ, we avoid free-form generation and score the labels
A--E directly from next-token logits. This removes the need for fragile answer
extraction and gives each model a reusable score vector for fusion. For OpenQA,
we use concise deterministic generation followed by cleanup that removes
reasoning traces, answer prefixes, XML-like tags, and extra whitespace. We then
combine cleaned answers from several Qwen-family models using development COMET
scores.

We also tested two common extensions, LoRA fine-tuning and OCR prompt
injection, but both reduced performance in our setting. The final system
therefore stays deliberately simple: pretrained VLMs, deterministic inference,
image enhancement, score-level ensembling for MCQ, and answer-level combination
for OpenQA. Most experiments were run on a single NVIDIA RTX PRO 6000 
GPU with approximately 95\,GB of memory.

Our official submissions ranked \textbf{third in Visual MCQ} with
\textbf{0.7108} accuracy and \textbf{first in Visual OpenQA} with
\textbf{0.6488} COMET. Overall, the results show that careful inference design
and output normalization can be highly competitive without task-specific model
training.

The main contributions of this paper are:
\begin{itemize}
  \item direct candidate label scoring for Visual MCQ, avoiding free-form answer extraction
  \item score fusion and voting over complementary VLM runs;
  \item concise Visual OpenQA generation with cleanup and weighted answer
        combination;
  \item ablations showing that LoRA fine-tuning and raw OCR prompt injection
        hurt performance in this setting.
\end{itemize}

\section{Related Work}

\subsection{Multilingual Exam-Style Visual Question Answering}

Multimodal exam question answering extends standard visual question answering
by combining visual text reading, diagram interpretation, mathematical notation,
tables, and domain knowledge. The ImageCLEF Multimodal Reasoning task follows
this direction. Its first edition was based on EXAMS-V, a multilingual benchmark
of exam questions containing diagrams, formulas, tables, and dense visual
text~\cite{ImageCLEFMultimodalReasoningOverview2025,EXAMSV2025}. The 2026
edition extends this setting with Visual MCQ and Visual OpenQA, where systems
must both reason over the image and satisfy strict output
formats~\cite{ImageCLEFMultimodalReasoningTaskOverview2026}.

Related benchmarks study individual parts of this problem. DocVQA focuses on
question answering over document images, where OCR quality and layout
understanding are central~\cite{Mathew2021DocVQA}. ChartQA evaluates chart
reading and numerical reasoning~\cite{Masry2022ChartQA}. ScienceQA studies
multimodal science questions~\cite{Lu2022ScienceQA}, while MMMU targets
expert-level academic reasoning across disciplines~\cite{Yue2024MMMU}.
ImageCLEF-MR combines these challenges in a multilingual shared-task setting,
making performance sensitive not only to visual reasoning but also to answer
language, formatting, and evaluation protocol.

\subsection{Vision-Language Models for Visual and Document Reasoning}

Vision-language models have moved from image-text alignment toward
instruction-following multimodal reasoning. CLIP introduced strong zero-shot
image-text matching through contrastive pretraining~\cite{Radford2021CLIP}.
BLIP-2 connected frozen visual encoders to large language models through a
query transformer~\cite{Li2023BLIP2}, while InstructBLIP and LLaVA showed the
importance of visual instruction tuning for general-purpose multimodal
assistants~\cite{Dai2023InstructBLIP,Liu2023LLaVA}.

Recent open VLMs further improve high-resolution image understanding,
multilingual processing, and document-like visual inputs. Qwen-VL and Qwen3-VL
are especially relevant because they provide strong visual instruction following
and visual text understanding~\cite{QwenVL2025,Qwen3VL2025}. InternVL also
targets broad multimodal reasoning with open checkpoints~\cite{InternVL2025}.
However, exam-style evaluation exposes a practical gap: even when a model can
understand the image, it can still lose accuracy through verbose answers,
unstable formatting, or failure to follow the required output structure.

\subsection{Output Control and Test-Time Aggregation}

Multiple-choice evaluation of language models often scores candidate
continuations instead of generating free-form answers, as in MMLU-style
evaluation~\cite{Hendrycks2021MMLU}. This avoids the ambiguity of extracting a
final label from a long explanation. The same idea is well suited to Visual MCQ:
because the answer set is fixed, the model can score labels A--E directly and
produce comparable score vectors for fusion.

Test-time aggregation is also widely used to improve robustness. In language
reasoning, self-consistency samples multiple reasoning paths and selects the
most consistent answer~\cite{Wang2023SelfConsistency}. In shared-task systems,
aggregation can combine different models, prompts, or checkpoints. The
ImageCLEF 2025 MSA system also showed that strict answer prompts reduce verbose
overflow responses in multimodal exam settings~\cite{Ahmed2025MSA}. These
findings motivate treating answer control, candidate scoring, and ensembling as
core parts of the system rather than post-processing details.

\section{Task and Data}

The ImageCLEF-MR 2026 task contains two visual question answering subtasks over
multilingual exam-style images. Each example provides a question image and a
question identifier. The images may contain printed text, diagrams, tables,
charts, formulas, option labels, and units, so the models must combine visual
text reading with scientific and educational reasoning.

The Visual MCQ subtask requires a JSON list with the question identifier and one
answer key from A--E. We used EXAMS-V~\cite{EXAMSV2025} for development and
model selection; the development split used in our experiments contained 3,565
labeled examples. The ImageCLEF-MR 2026 Visual MCQ test split contains 1,117
examples across English, Bulgarian, Chinese, Croatian, Italian, and Serbian.
The official metric is accuracy, reported overall and by language.

The Visual OpenQA subtask uses the same type of visual input but expects a short
free-form answer string rather than a fixed option label. The 2026 OpenQA data
contains 528 labeled training examples and 439 test examples. Systems are
evaluated with COMET~\cite{Rei2020COMET}, BLEU, ROUGE-L, and METEOR; the
leaderboard is ranked by COMET.

\begin{figure}[htbp]
    \centering
    \begin{minipage}{0.48\linewidth}
        \centering
        \includegraphics[width=\linewidth]{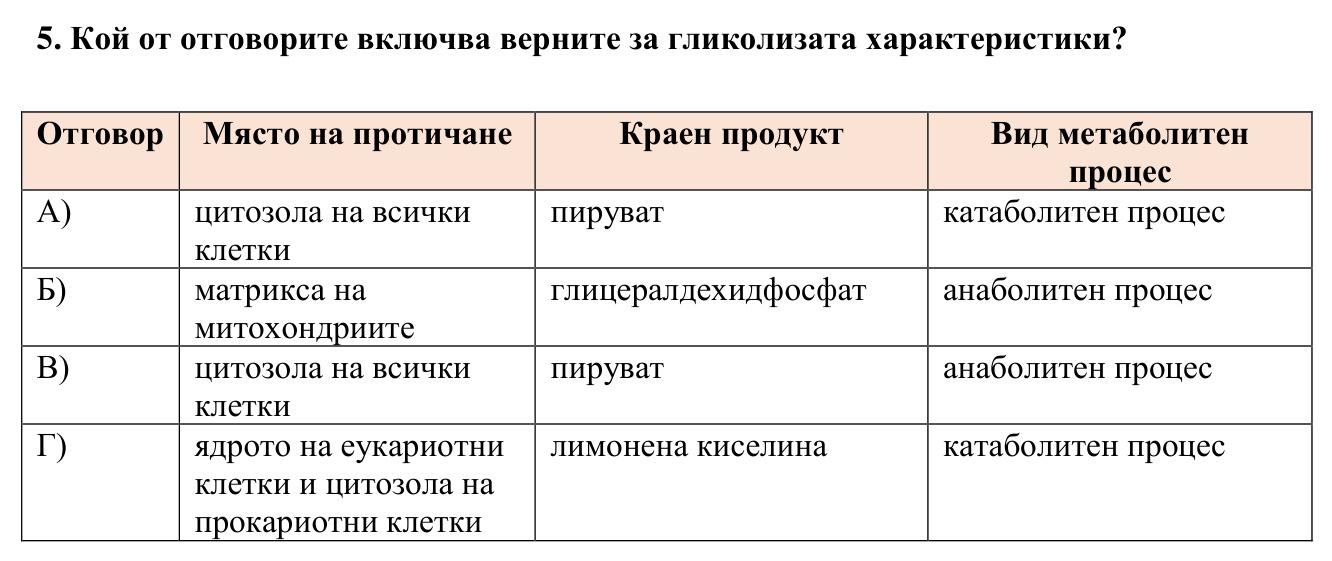}
        \small (a) Visual MCQ
    \end{minipage}
    \hfill
    \begin{minipage}{0.48\linewidth}
        \centering
        \includegraphics[width=\linewidth]{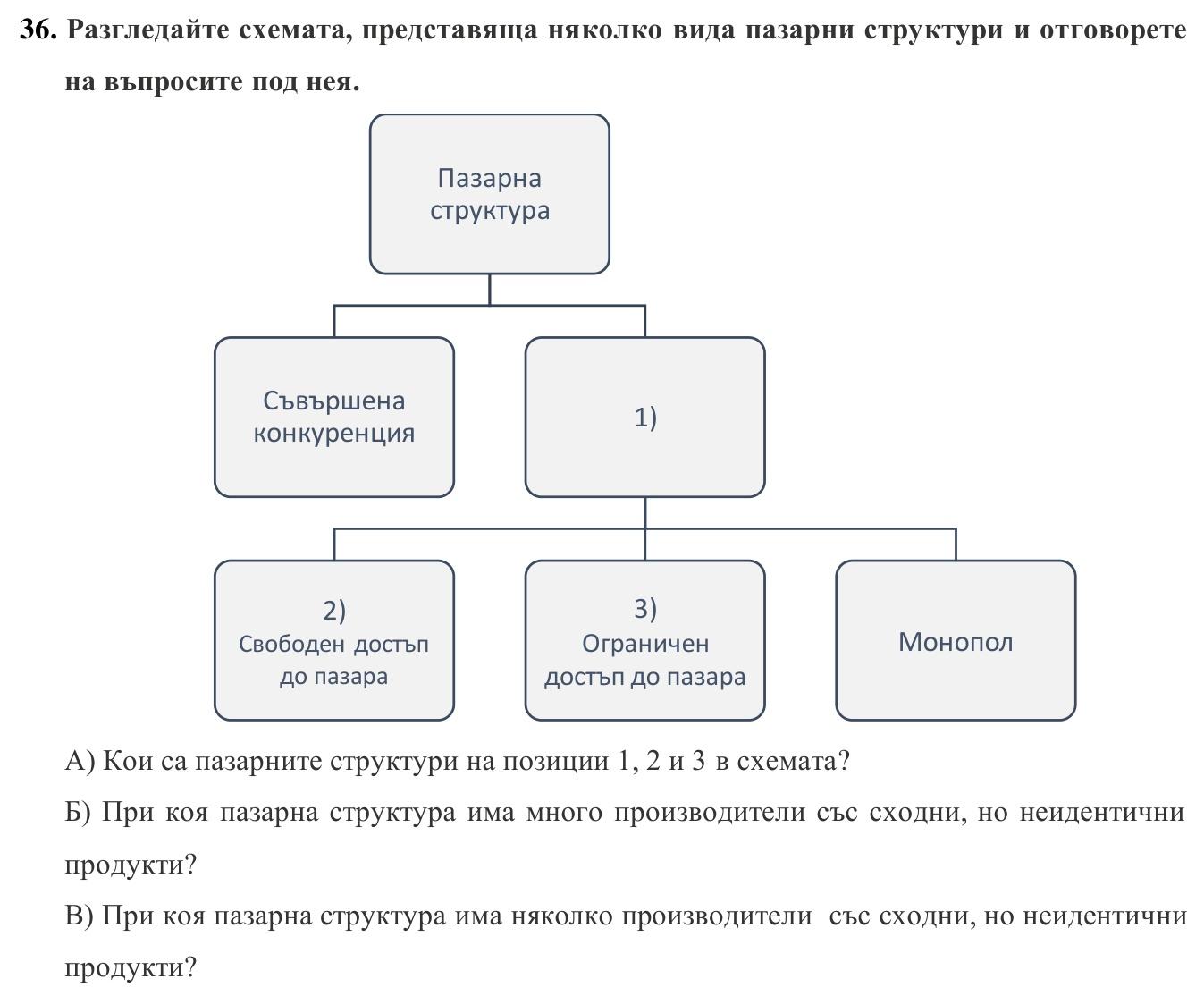}
        \small (b) Visual OpenQA
    \end{minipage}
    \caption{Example ImageCLEF-MR 2026 samples. The MCQ example shows a table-based Bulgarian biology question with answer options, while the OpenQA example shows a diagram-based Bulgarian economics question requiring short free-form answers.}
    \label{fig:dataset-sample}
\end{figure}

\section{System Overview}

Our system follows a shared front end with task-specific decision logic. Images
are loaded through PIL, kept in a standard RGB representation for model
compatibility, resized to a controlled longest side, and lightly enhanced for
contrast and sharpness. The associated metadata, question text, and answer
options are then inserted into a task-specific prompt.

For Visual MCQ, we cast answer selection as candidate scoring rather than
free-form generation. The model's next-token logits score the candidate labels
A--E whenever available, producing both a predicted label and a reusable score
vector for fusion. Runs without usable logits are treated as voters.

For Visual OpenQA, where no fixed label set exists, we use concise deterministic
generation and post-processing to remove reasoning traces, XML-like tags, answer
prefixes, and duplicate whitespace. Thus, both subtasks follow the same
principle: restrict the model output to the format required by the evaluator.

Fine tuned and OCR-augmented variants were evaluated within the same framework, but they are reported as ablations because they did not improve the final submissions.

Across final runs, we used deterministic decoding, enhanced RGB images, batch
size one, next-token scoring for MCQ when available, and a maximum of 192 new
tokens for OpenQA. Intermediate JSON/JSONL files were written after each example
so long jobs could be resumed safely.

All scripts were implemented in Python using Hugging Face \texttt{datasets}, \texttt{transformers}, PyTorch, and Unsloth for Qwen3.6 MoE inference. Most final experiments were run on a single NVIDIA RTX PRO 6000 Blackwell GPU with about 95GB of memory; smaller smoke tests were run on an NVIDIA L4 with about 22GB of memory.

\subsection{Implementation Settings}

Table~\ref{tab:implementation-settings} summarizes the main inference and diagnostic training settings. Official submissions used inference only. LoRA and QLoRA runs were post-hoc diagnostic experiments and were not included in the final submitted systems.

\begin{table*}[htbp]
\caption{Main implementation settings used for reproducibility.}
\label{tab:implementation-settings}
\small
\centering
\begin{tabular}{@{}p{0.18\linewidth}p{0.72\linewidth}@{}}
\toprule
Component & Setting \\
\midrule
MCQ inference & Batch size 1; enhanced images with 768-pixel longest side; strict \texttt{ANSWER:} prefill; greedy generation for generation-only runs; next-token A--E scoring when logits were available. \\
OpenQA inference & Batch size 1; enhanced images; deterministic decoding without sampling; maximum 192 new tokens; answer cleanup before metric computation and ensembling. \\
Qwen3-VL-8B LoRA & EXAMS-V train split; seed 42; 300-step diagnostic run; 4-bit QLoRA; batch size 1 with 8-step gradient accumulation; learning rate \(1\times10^{-4}\); cosine schedule; 3\% warmup; LoRA \(r=16\), \(\alpha=32\), dropout 0.05; target modules \texttt{q\_proj}, \texttt{k\_proj}, \texttt{v\_proj}, \texttt{o\_proj}, \texttt{gate\_proj}, \texttt{up\_proj}, and \texttt{down\_proj}. \\
Qwen3.6-35B QLoRA & EXAMS-V and post-release ImageCLEF MCQ diagnostic runs; seed 42; 4-bit Unsloth training; batch size 1 with 8-step gradient accumulation; learning rate \(3\times10^{-5}\); cosine schedule; 3\% warmup; LoRA \(r=16\), \(\alpha=16\), dropout 0.0; all linear modules targeted. \\
Trainable weights & Base model weights were frozen in the LoRA/QLoRA runs; only adapter weights were updated. These adapters were evaluated as ablations because they did not improve the final systems. \\
\bottomrule
\end{tabular}
\end{table*}

\FloatBarrier
\section{Visual MCQ Method}

\subsection{Image Preparation}

Each MCQ image was loaded in RGB, resized to a 768-pixel longest side, and
lightly enhanced for contrast and sharpness. This setting balanced readability
with throughput across 1,117 test examples and several model variants.

\subsection{Prompting and Candidate Label Scoring}

The MCQ prompt was intentionally strict. It asked the model to inspect the question image, answer options, labels, formulas, units, tables, and diagrams, but to return only one uppercase label from A--E. For models that used generation, we prefixed the answer with \texttt{ANSWER:} to reduce formatting variation. For models that exposed logits, the prompt served as context for direct candidate label scoring rather than as a request for a long explanation.

In the MCQ subtask, a generated explanation is unnecessary and can be harmful. We therefore score the five answer labels directly whenever logits are available. In implementation, the image and prompt are first encoded as a chat-style input and the answer prefix \texttt{ANSWER:} is appended. We then run a single forward pass, read the logits at the next-token position, and collect the logits associated with the candidate labels A--E. If a tokenizer exposes multiple equivalent label forms, such as a bare letter and a whitespace-prefixed letter, we keep the larger logit for that label. The raw score dictionary is stored for every example so it can be inspected and reused by the ensemble.

Given image \(I\), prompt \(p\), and candidate answer \(c \in \{A,B,C,D,E\}\), the model produces a score \(s(c \mid I,p)\) from the next-token logits. The prediction is:
\begin{equation}
  \hat{y} = \argmax_{c \in \{A,B,C,D,E\}} s(c \mid I,p)\;.  
\end{equation}
This avoids regular expression extraction from long generations and produces comparable per-answer score dictionaries for ensembling. Generation-only runs were still cleaned with the same rule: strip special tokens and answer prefixes, extract the first valid uppercase A--E label, and validate the final official JSON schema.

\subsection{Model Selection}

Our most successful single-model MCQ setting was direct candidate scoring without fine-tuning or OCR injection. Table~\ref{tab:mcq successful runs} lists the single-model and diagnostic MCQ results that we evaluated on the released test labels. Qwen3.6-35B-A3B base and  Qwen3.6-27B abliterated were the strongest individual scorers and made partially different mistakes, which motivated score-level merging. The Qwen3-VL-8B Thinking model was also evaluated both directly and with LoRA 300 fine-tuning; however, the fine-tuned run worsened the results compared to the direct run and remained below the strongest base scorers.

\begin{table}[htbp]
\caption{Single model Visual MCQ results. Accuracies are post-release local evaluations on the 1,117 Visual MCQ test examples using the leaderboard-compatible multi-answer interpretation.}
\label{tab:mcq successful runs}
\small
\begin{tabular}{@{}p{0.58\linewidth}cc@{}}
\toprule
Model & Size & Accuracy \\
\midrule
Qwen3.6-35B-A3B base & 36.1B/A3B & \textbf{0.6849} \\
 Qwen3.6-27B abliterated & 27B & 0.6822 \\
Qwen2.5-VL-32B & 32B & 0.6052 \\
Qwen3.6-35B-A3B LoRA epoch 1 & 36.1B/A3B & 0.5801 \\
Qwen3-VL-8B Thinking direct & 8B & 0.5774 \\
Qwen3-VL-8B Thinking LoRA 300 & 8B & 0.5649 \\
 Qwen3.6-27B with OCR prompt & 27B & 0.5542 \\
Qwen2.5-VL-7B multilingual enhanced & 7B & 0.5479 \\
Qwen2.5-VL-7B weak stage enhanced & 7B & 0.5434 \\
Qwen2.5-VL-7B LoRA & 7B & 0.5407 \\
Qwen3-VL-8B Thinking with OCR prompt & 8B & 0.5103 \\
\bottomrule
\end{tabular}
\end{table}

\subsection{Fine-Tuning and OCR Ablation}

After identifying strong zero-shot candidate scorers, we tested two natural ways to improve them: supervised adaptation and external OCR. Both were useful experiments, but neither improved the final MCQ system. For fine-tuning, we trained and evaluated two LoRA-style tracks: a Qwen3.6-35B-A3B QLoRA checkpoint and a Qwen3-VL-8B Thinking LoRA 300 checkpoint. The Qwen3.6 LoRA checkpoint was weaker than the base model on both EXAMS-V development evaluation and the released ImageCLEF-MR 2026 MCQ test labels. For Qwen3-VL-8B Thinking, the direct run scored 0.5774 and the LoRA 300 run scored 0.5649, so the adapter also reduced accuracy. This suggests that the available supervised data was too small and too heterogeneous for stable adaptation, and that many errors came from visual reading, formulas, and layout rather than missing language model knowledge.

For OCR, we used OCR.space engine 2 and retried failed calls until the final MCQ OCR file covered all 1,117 test examples. We appended OCR text inside an \texttt{<ocr\_text>} block and instructed the model to use it only as supporting evidence. This also hurt. The raw OCR transcript often contained duplicated fragments, broken formulas, partial option labels, and reading order that disagreed with the visual layout. As a result, it contaminated the prompt and shifted probability mass toward wrong labels for models that already had visual text reading ability.

\begin{center}
\refstepcounter{table}\label{tab:mcq negative ablation}
\small
\textbf{Table~\thetable: Negative MCQ Ablations.} Scores use the leaderboard-compatible multi-answer interpretation when evaluated on the released ImageCLEF-MR 2026 MCQ test labels.

\begin{tabular}{@{}p{0.56\linewidth}cc@{}}
\toprule
Experiment & Reference & Result \\
\midrule
Qwen3.6 QLoRA on EXAMS-V & 0.5910 & 0.4856 \\
Qwen3.6 QLoRA on MCQ test & 0.6849 & 0.5801 \\
Qwen3-VL-8B direct \(\rightarrow\) LoRA 300 & 0.5649 & 0.5774  \\
 Qwen3.6-27B + OCR  & 0.6822 & 0.5542 \\
Qwen3-VL-8B + OCR  & 0.5774 & 0.5103 \\
\bottomrule
\end{tabular}
\end{center}

The Qwen3.6 base + OCR run failed during loading on an A100 40GB GPU because some modules were dispatched to CPU or disk. We therefore kept fine-tuned and OCR-augmented runs as diagnostic ablations rather than final system components. The final MCQ system instead preserved the pretrained models' direct visual reasoning behavior and combined their raw label scores.

\subsection{Score-Level Combination}

When raw A--E score dictionaries were available, we used a weighted score fusion, i.\,e.,
\begin{equation}
  S(c) = \sum_i w_i s_i(c)\;.  
\end{equation}
The answer with maximum \(S(c)\) is selected. When only final labels were available, we used majority voting with deterministic tie priority. The two MCQ combination rules are formalized in Appendix~\ref{app:ensemble-algorithms}, Algorithm~1 for weighted score fusion and Algorithm~2 for majority voting with tie priority. Table~\ref{tab:mcq-runs} summarizes these ensemble and merge experiments in one place. The strongest post-release local fusion combined Qwen3.6-27B abliterated and Qwen3.6-35B-A3B base logits with weights 2:1, scoring 0.7126. Our official submitted ensemble, \texttt{mcq\_ensemble\_weighted\_vote}, scored 0.7108.
\begin{figure}
    \centering
    \includegraphics[width=0.9\linewidth]{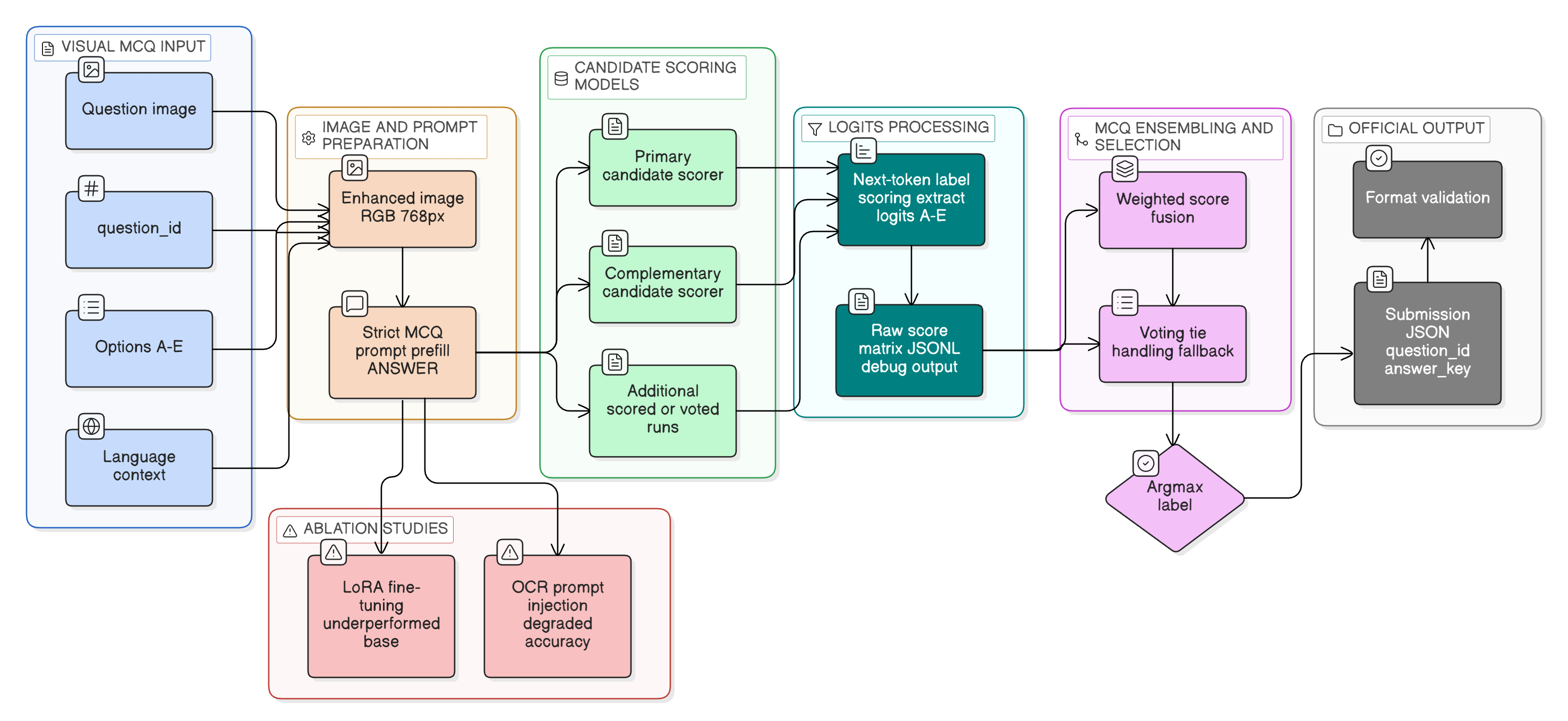}
    \caption{Visual MCQ inference workflow. Each sample is enhanced, paired with a strict answer-label prompt, processed by candidate-scoring VLMs, converted into raw A--E logit scores, and merged through score fusion or voting before official JSON validation. Red dashed paths show ablations that were tested but excluded from the final system because they reduced accuracy.}
    \label{fig:mcq-pipeline}
\end{figure}

Figure~\ref{fig:mcq-pipeline} summarizes the MCQ path as an engineering pipeline rather than a single model call. The important design choice is that each model produces a reusable score vector over A--E, not just a final text answer. This made the strongest direct-scoring runs directly comparable and allowed weighted fusion, while the OCR and LoRA branches were recorded as negative ablations instead of being included in the final submission.

\begin{table*}[htbp]
\caption{Visual MCQ ensemble and score merge runs. The best local fusion is shown in bold; the official submitted run is shown in the second row.}
\label{tab:mcq-runs}
\small
\centering
\setlength{\tabcolsep}{4pt}
\renewcommand{\arraystretch}{0.7}
\begin{tabular}{@{}>{\raggedright\arraybackslash}p{0.31\linewidth}>{\raggedright\arraybackslash}p{0.18\linewidth}>{\raggedright\arraybackslash}p{0.36\linewidth}c@{}}
\toprule
Run / models & Merge method & Notes & Accuracy \\
\midrule
\textbf{ Qwen3.6-27B + Qwen3.6-35B-A3B base} & \textbf{Weighted score fusion} & \textbf{Best post-release local merge; raw A--E logits with weights 2.0 and 1.0} & \textbf{0.7126} \\
Official MCQ ensemble & Weighted voting / score fusion & Submitted run; combined Huihui-Qwen3.6-27B, Qwen3.6-35B-A3B, Qwen2.5-VL-32B, and older Qwen-family runs & 0.7108 \\
\addlinespace[2pt]
Qwen3.6-35B-A3B base + Qwen3.6 LoRA & Weighted score fusion & Base favored weights 3.0 and 1.0 & 0.6885 \\
Qwen3.6-35B-A3B base + Qwen3.6 LoRA & Confidence router & Routed 759 examples to the base model and 358 to LoRA & 0.6777 \\
Qwen3.6-35B-A3B base + Qwen3.6 LoRA & Weighted score fusion & Equal 1.0 and 1.0 weights & 0.6688 \\
Qwen3.6-35B-A3B base + Qwen3.6 LoRA & Weighted score fusion & LoRA favored weights 1.0 and 2.0 & 0.6509 \\
\addlinespace[2pt]
Qwen3.6 LoRA + Qwen3.6 base + old Qwen vote & Majority vote & Tie priority favored newer Qwen3.6 runs & 0.6455 \\
Old Qwen vote + Qwen3.6 LoRA + Qwen3.6 base & Majority vote & Alternate tie priority & 0.6401 \\
Three Qwen2.5-VL-7B variants & Majority vote & LoRA, multilingual enhanced, and weak stage variants & 0.5452 \\
\bottomrule
\end{tabular}
\end{table*}

The ensemble results show that merging the two strongest complementary base predictors was more useful than simply adding many weaker systems. In contrast, OCR-augmented runs and weak Qwen2.5-VL-7B variants were useful as ablations and historical baselines, but they did not improve the final score when added without additional filtering.

\FloatBarrier
\section{Visual OpenQA Method}

The Visual OpenQA subtask requires a short free-form answer rather than a
selection from predefined options. This makes the task considerably more
sensitive to verbosity, formatting errors, and reasoning traces. A model that
correctly interprets the visual evidence can still receive a low score if it
produces an explanation instead of a direct answer, generates text in the wrong
language, or copies irrelevant fragments from the image. We therefore treated
OpenQA as a controlled short answer generation problem with four sequential
stages: image preparation, concise prompting with deterministic decoding, output
normalization, and answer-level combination across models.

\subsection{Image Preparation}

For OpenQA enhanced-image runs, each image was loaded in RGB format, resized to a
controlled longest side, and adjusted for contrast and sharpness. The RGB
conversion was only a defensive normalization step. The Qwen3.6 and generic
OpenQA runners used a longest side of 1000 pixels, while Qwen-VL runners used
1600 pixels unless overridden. These higher resolutions helped preserve fine details such as numbers, labels, chart ticks, formulas, and unit symbols.

We also collected OCR.space outputs for all 528 training examples and all 439
test examples after retries. The final training OCR file contained one non-empty
OCR record for every example. However, OCR was used only as diagnostic evidence.
Raw OCR often introduced duplicated fragments, broken formulas, and incorrect
reading order. Since OpenQA is sensitive to exact wording, injecting noisy OCR
risked irrelevant copying and overlong answers. The official OpenQA submission
therefore relied entirely on the image-based pipeline.

\subsection{Prompting and Deterministic Decoding}

The OpenQA prompt was intentionally compact. Long instructions tended to
encourage the model to explain its reasoning, translate the question, or
reproduce large portions of the image text. The final prompt instead imposed
four hard constraints: (i)~answer in the question language when possible, (ii)~keep the
output concise, (iii)~preserve exact numbers and units, and (iv)~suppress all reasoning
traces. The full template is shown in Appendix~\ref{app:prompts}.

We used greedy (deterministic) decoding with no sampling and capped generation
at 192 new tokens. The cap is generous enough to accommodate multi part answers,
but the prompt and post-processing were designed to make most outputs far shorter
in practice. Deterministic decoding also reduced run-to-run variance, which
mattered for reproducible comparison across models.

Unlike the MCQ pipeline, where the final decision is based on candidate label logits, OpenQA has no fixed answer set. Consequently, there is no equivalent to next-token scoring over A--E. The system instead relies on controlled generation followed by cleanup.

\subsection{Model Selection}

We evaluated several open or locally runnable VLMs on the 528 labeled OpenQA
training examples. Results are summarized in Table~\ref{tab:openqa dev}.
Qwen3-VL-32B Thinking achieved the best single-model COMET score (0.6270).
Qwen2.5-VL-32B Instruct was the next strongest run at 0.6125 COMET, slightly
ahead of Qwen3-VL-8B Thinking (0.6093), which was considerably more practical
to run. Qwen3.6-35B-A3B was weaker by COMET but produced competitive METEOR
and ran stably under the Unsloth 4 bit inference setup despite its larger
parameter count.
InternVL3, Aya Vision, Phi-4 Reasoning Vision, and Mistral Small Vision served
as useful baselines but underperformed the strongest Qwen-family runs, partly
because they were more prone to verbose outputs or inconsistent language
adherence.

These observations directly shaped the final submission: we retained the
Qwen-family models and excluded models that were too slow, unstable under
available memory, or structurally prone to generating long reasoning traces. The
final OpenQA submission ensemble therefore includes Qwen3-VL-32B Thinking,
Qwen3-VL-8B Thinking, Qwen2.5-VL-32B Instruct, and Qwen3.6-35B-A3B.

\begin{table}[htbp]
\caption{Visual OpenQA single-model development results on the 528 labeled
training examples, ranked by COMET.}
\label{tab:openqa dev}
\small
\begin{tabular}{@{}lccccc@{}}
\toprule
Model & Size & BLEU & ROUGE L & METEOR & COMET \\
\midrule
Qwen3-VL-32B Thinking      & 32B         & 0.1076 & 0.1694 & 0.1514 & \textbf{0.6270} \\
Qwen2.5-VL-32B Instruct    & 32B         & 0.0973 & 0.1689 & 0.1488 & 0.6125 \\
Qwen3-VL-8B Thinking       & 8B          & 0.0848 & 0.1598 & 0.1274 & 0.6093 \\
Qwen3.6-35B-A3B base       & 36.1B/A3B   & 0.0657 & 0.1418 & 0.1456 & 0.5663 \\
InternVL3-8B-hf            & 8B          & 0.0326 & 0.1126 & 0.1030 & 0.4681 \\
Aya Vision 32B             & 32B         & 0.0194 & 0.0664 & 0.0576 & 0.4667 \\
Phi-4 Reasoning Vision     & 15B         & 0.0268 & 0.0815 & 0.0554 & 0.4344 \\
Mistral Small FP8          & 24B         & 0.0040 & 0.0200 & 0.0064 & 0.4248 \\
InternVL3.5 38B            & 38B         & 0.0076 & 0.0369 & 0.0401 & 0.4150 \\
\bottomrule
\end{tabular}
\end{table}

\subsection{OCR Ablation}

To test whether external OCR could improve OpenQA, we reran the two strongest
Qwen3-VL models on the labeled training split with the final OCR file injected
into the prompt. The OCR aware prompt explicitly stated that the image remained
the primary source of truth and that OCR should only be used as supporting
evidence for small or blurry text. This was intended to help with diagrams,
tables, formulas, and chart labels without encouraging the model to copy the OCR
transcript.

Table~\ref{tab:openqa ocr ablation} shows that OCR did not improve either
model. Qwen3-VL-32B Thinking dropped from 0.6270 to 0.6078 COMET, while
Qwen3-VL-8B Thinking dropped slightly from 0.6093 to 0.6043. The 32B model did
gain higher BLEU and ROUGE L than some previous non OCR runs, but COMET, the
official ranking metric, decreased. We therefore excluded OCR prompt injection
from the final OpenQA submission.

\begin{table}[htbp]
\caption{OpenQA OCR ablation on the 528 labeled training examples. OCR was
injected into the prompt using the final complete train OCR file.}
\label{tab:openqa ocr ablation}
\small
\begin{tabular}{@{}lccccc@{}}
\toprule
Model & OCR & BLEU & ROUGE L & METEOR & COMET \\
\midrule
Qwen3-VL-32B Thinking & \mycross & 0.1076 & 0.1694 & 0.1514 & 0.6270 \\
Qwen3-VL-32B Thinking & \mycheck    & 0.1040 & 0.1698 & 0.1264 & 0.6078 \\
Qwen3-VL-8B Thinking  & \mycross & 0.0848 & 0.1598 & 0.1274 & 0.6093 \\
Qwen3-VL-8B Thinking  & \mycheck    & 0.0857 & 0.1496 & 0.1110 & 0.6043 \\
\bottomrule
\end{tabular}
\end{table}

The most likely reason is that strong VLMs already read much of the image text
directly, while OCR introduces a second, noisier textual view of the same image.
When this text contains duplicated fragments, broken symbols, or layout order
that differs from the figure, the model may over trust the OCR and generate
longer or less focused answers. This made OCR useful for analysis and possible
future filtering, but not as a direct prompt input in our final OpenQA system.

\subsection{Answer Cleaning}

Post processing was an essential step. 
Thinking-style models routinely produced \texttt{<think>} blocks, XML-like tags, reasoning traces, answer prefixes such as ``Answer:'', and explanatory lead-ins, even when the prompt requested only the final answer. We removed these artifacts with deterministic rules: delete complete \texttt{<think>...\allowbreak</think>} spans, strip residual XML-like tags and model special tokens, remove leading answer prefixes, collapse whitespace, drop blank strings, and remove exactly repeated answer fragments.

This cleaning step has a direct impact on all four official metrics. Extraneous
text reduces BLEU, ROUGE L, and METEOR by diluting $n$ gram overlap with the
reference. It also lowers COMET when the additional text shifts the semantic
focus of the answer away from the correct information. Aggressive cleanup was
therefore not optional: it was a prerequisite for competitive metric scores.

\subsection{Answer Level Ensembling}
\label{sec:openqa-ensemble}

\paragraph{Input runs.}
The OpenQA ensemble was implemented in \texttt{scripts/ensemble\_openqa.py} and
starts after the answer-cleaning stage described above. At this point, each
model provides a cleaned answer-list for each question, so the ensemble focuses
only on selecting among competing answer strings. We combine official-format
prediction files from five models: Qwen3-VL-32B-Thinking,
Qwen3-VL-8B-Thinking, Qwen2.5-VL-32B-Instruct, Qwen3.6-35B-A3B, and
InternVL3-8B-hf.

\paragraph{Model weighting.}
Each model is assigned a scalar weight derived from its development-set COMET
score. To sharpen the distribution and suppress the influence of weaker models,
we apply a softmax over \(10 \times\) the raw score:
\begin{equation}
  w_m = \frac{\exp(10s_m)}{\sum_j \exp(10s_j)}\;.
  \label{eq:softmax-weights}
\end{equation}
This yields approximate weights of 29.0\,\% for Qwen3-VL-32B, 25.1\,\% for
Qwen2.5-VL-32B, 24.3\,\% for Qwen3-VL-8B, 15.8\,\% for Qwen3.6-35B-A3B, and
5.9\,\% for InternVL3-8B. The temperature factor of 10 concentrates probability on stronger models while still allowing lower-ranked models to exert meaningful influence when they converge on an answer the top model misses.

\paragraph{Combination strategies.}
We submitted three variants, formalized in
Appendix~\ref{app:ensemble-algorithms}. \breakableensemblepositionweighted{}
clusters answers at each position using Jaccard similarity \(\geq 0.35\), then
selects the cluster with the largest total model weight. The answer count is
chosen by weighted majority vote, preferring the top two models when they agree.
\breakableensembletoptwounion{} is more conservative: it keeps Qwen3-VL-32B as
the backbone and replaces an answer only when Qwen3-VL-8B disagrees and
Qwen2.5-VL-32B supports the 8B answer. \breakableensembleweightedunion{} is
the recall-oriented version: it pools unique answers from all models, merges
similar strings, and returns the top-\(N^*\) weighted clusters. Answer-level
combination improved over the best single-model: the best official OpenQA COMET
score was \textbf{0.6488}, compared with 0.6270 for Qwen3-VL-32B-Thinking
alone.
\begin{figure}
    \centering
    \includegraphics[width=0.9\linewidth]{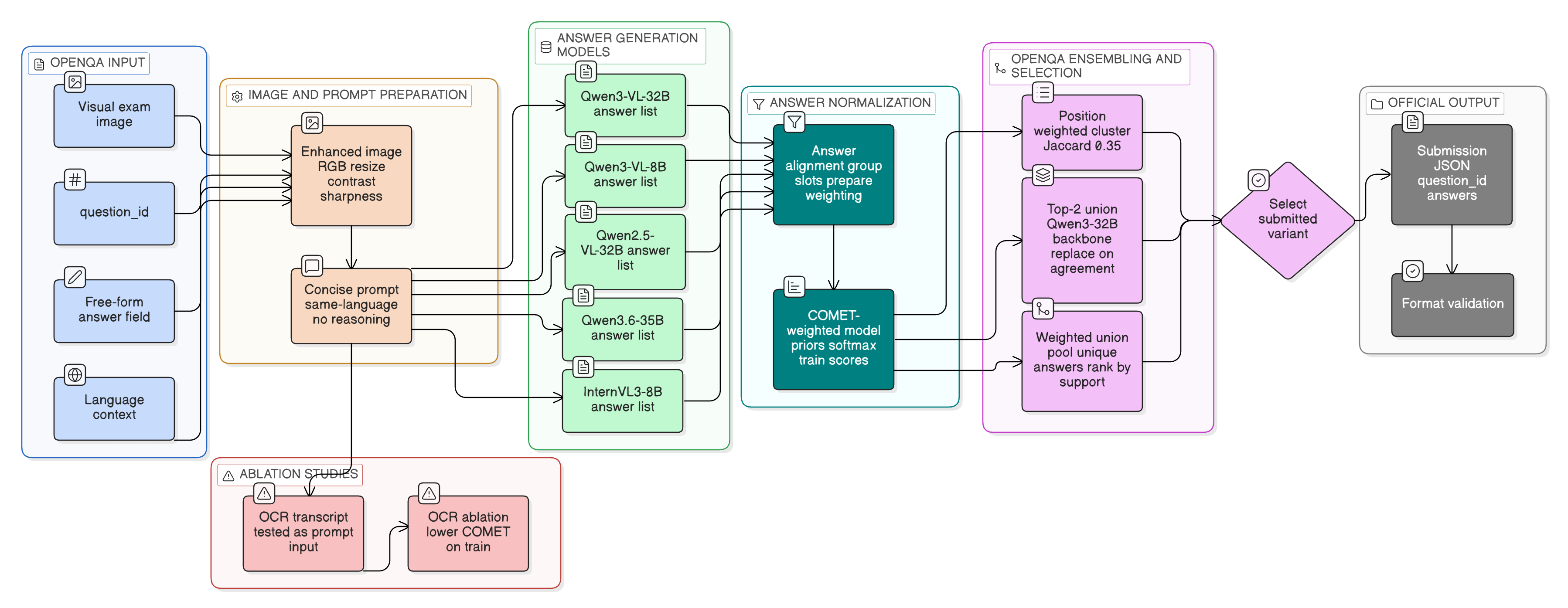}
\caption{Visual OpenQA pipeline. The image and metadata are standardized and
paired with a concise no-reasoning prompt. Five VLMs generate answer lists,
which are aligned and weighted using development-set COMET scores. The system
then produces three answer-level ensemble variants: position-weighted
clustering, conservative top-2 union, and weighted union. The selected variant
is written as the official \texttt{question\_id}/\texttt{answers} JSON. The OCR
branch shows an ablation that was tested but excluded because it reduced COMET.}
\label{fig:openqa-pipeline}
\end{figure}

Figure~\ref{fig:openqa-pipeline} shows the OpenQA path as an answer-bank
workflow rather than a label-scoring workflow. The key difference from MCQ is
that there is no fixed A--E label set, so the system combines generated answer
lists only after model inference and answer cleanup.

\FloatBarrier
\section{Official Submitted Runs and Results}
\label{sec:official results}

The official submissions use the same two decision paths described above. The
Visual MCQ submission, \texttt{mcq\_ensemble\_weighted\_vote}, is a score/vote
ensemble over Huihui-Qwen3.6-27B, Qwen3.6-35B-A3B, Qwen2.5-VL-32B, and earlier
Qwen-family runs. The strongest Visual OpenQA submission uses answer-level
combination over Qwen3-VL-32B, Qwen3-VL-8B, Qwen2.5-VL-32B, Qwen3.6-35B-A3B,
and InternVL3-8B.

Table~\ref{tab:mcq official} shows the official Visual MCQ leaderboard excerpt. Our submission ranked third overall with 0.7108 accuracy. The strongest language score for our system was Italian, while Chinese remained the hardest language.

\begin{table}[!htbp]
\caption{Official Visual MCQ leaderboard excerpt.}
\label{tab:mcq official}
\small
\begin{tabular}{@{}lccccccc@{}}
\toprule
Team & Overall & EN & BG & ZH & HR & IT & SR \\
\midrule
spirosbax & \textbf{0.8406} & \textbf{0.8560} & \textbf{0.8909} & \textbf{0.7807} & \textbf{0.8772} & \textbf{0.9074} & \textbf{0.8704} \\
DS@GT & 0.7986 & 0.8520 & 0.8636 & 0.6725 & 0.8596 & 0.8704 & 0.8333 \\
mohamedbasem (ours) & 0.7108 & 0.7480 & 0.6909 & 0.6345 & 0.7368 & 0.8148 & 0.7593 \\
\bottomrule
\end{tabular}
\end{table}

Table~\ref{tab:openqa official} shows the official Visual OpenQA leaderboard excerpt. Our submission ranked first by COMET.

\begin{table}[!htbp]
\caption{Official Visual OpenQA leaderboard excerpt.}
\label{tab:openqa official}
\small
\begin{tabular}{@{}lcccc@{}}
\toprule
Team & COMET & BLEU & ROUGE L & METEOR \\
\midrule
mohamedbasem (ours) & \textbf{0.6488} & \textbf{0.1391} & \textbf{0.2762} & 0.2383 \\
wangshou66 & 0.6366 & 0.1308 & 0.2717 & \textbf{0.2388} \\
uned martinez & 0.5938 & 0.0980 & 0.2452 & 0.1842 \\
liuzhe3085 & 0.5829 & 0.1556 & 0.3351 & 0.2295 \\
\bottomrule
\end{tabular}
\end{table}

\FloatBarrier
\section{Discussion}

The main lesson is that answer control was as important as model selection. For
Visual MCQ, direct A--E candidate scoring avoided unreliable extraction from
long generated explanations and produced comparable score vectors for
ensembling. For Visual OpenQA, concise prompting, deterministic decoding, and
strict cleanup reduced verbose outputs and reasoning traces. This is consistent
with prior ImageCLEF findings that strict prompts reduce overflow
responses~\cite{Ahmed2025MSA}.

The ensemble results show that complementarity mattered more than simply adding
more models. The strongest MCQ merge combined two strong scorers with different
error patterns, while the OpenQA ensemble benefited from selecting among
semantically similar answers from several Qwen-family models. Weak or noisy runs
only helped when filtered or weighted.

OCR and fine-tuning were mostly negative experiments. OCR often lost layout,
option order, formulas, or diagram structure, and reduced both MCQ accuracy and
OpenQA COMET. LoRA and QLoRA also failed to outperform the strongest direct
base-model runs. For this task, robust inference, image-level reading, and
careful formatting were more useful than small-data adaptation or raw OCR.

Runtime constraints also shaped the final system. Some large FP8 or MoE
checkpoints were slow or unstable, while stable Qwen-family runs could be
completed, validated, and ensembled. Evaluation details mattered as well: after
test labels were released, our MCQ run scored 0.6956 under strict equality and
0.7108 under the leaderboard-compatible multi-answer interpretation.

One limitation is that our observation about longer prompts encouraging longer
answers was based on development logs rather than an isolated prompt-length
study. We therefore treat it as an engineering observation, not as a causal
claim. A controlled ablation that varies only prompt length while holding model,
image resolution, decoding, and cleanup fixed would be useful future work.
\section{Conclusion}

We presented a practical system for both subtasks of the ImageCLEF 2026
Multimodal Reasoning challenge. The central finding is that output control is
a primary determinant of leaderboard performance in exam style multimodal
evaluation, on par with the choice of underlying vision-language model.

For Visual MCQ, reformulating prediction as direct candidate label scoring
eliminated the largest source of evaluation error in our early experiments.
Scoring next-token logits over A--E instead of extracting answers from generated
text produced reusable, comparable score vectors that supported principled
weighted fusion, achieving 0.7108 accuracy and third place on the official
leaderboard. For Visual OpenQA, concise prompting, deterministic decoding, and
aggressive post-processing combined with weighted answer-level combination across
Qwen-family models yielded 0.6488 COMET and first place   exceeding the best
single-model development score and confirming that late stage combination adds
meaningful value.

Our ablations produced two consistent negative results: LoRA fine-tuning on
small heterogeneous data and raw OCR prompt injection each degraded performance
relative to direct zero-shot scoring. Both findings suggest that current strong
VLMs handle visual text reading well enough that noisy external signals are
counterproductive rather than helpful.

The broader lesson is that the gap between a model's latent capability and its
evaluated performance can be large, and that closing this gap through inference
engineering is a tractable and underexplored complement to model scaling.

\section*{Declaration on Generative AI}
During the preparation of this work, the authors used generative AI tools,
including OpenAI ChatGPT/Codex, to support grammar checking, wording
improvement, LaTeX editing, code inspection, and organization of experimental
notes. The authors reviewed, verified, and edited the generated content as
needed and take full responsibility for the publication's content.

\appendix

\section{Prompt Templates}
\label{app:prompts}

The following prompts were used for the main image only runs and the OCR
diagnostic runs. In OCR runs, the external OCR transcript was appended to the
model input after the prompt as supporting evidence.

\subsection{Visual MCQ Prompt}
\begin{lstlisting}
You are solving a visual multiple choice exam question.

Read the image carefully in its original language. Use all visible evidence: question text, answer options, option labels, diagrams, charts, tables, legends, formulas, symbols, numbers, and units.

Compare every option against the visual evidence and subject knowledge. Choose the single best answer.

Output exactly one uppercase letter: A, B, C, D, or E.
Do not output words, punctuation, Markdown, XML tags, explanation, reasoning, or a sentence.
\end{lstlisting}

\subsection{Visual MCQ OCR Prompt}
\begin{lstlisting}
You are solving a visual multiple choice exam question.

Read the image carefully in its original language. Use all visible evidence: question text, answer options, option labels, diagrams, charts, tables, legends, formulas, symbols, numbers, and units.

OCR text may be provided after the image. Use OCR only as supporting evidence to recover small or blurry text. If OCR conflicts with the image, trust the image.

Compare every option against the visual evidence, OCR evidence, and subject knowledge. Choose the single best answer.

Output exactly one uppercase letter: A, B, C, D, or E.
Do not output words, punctuation, Markdown, XML tags, explanation, reasoning, or a sentence.
\end{lstlisting}

\subsection{Visual OpenQA Prompt}
\begin{lstlisting}
You are answering a visual open ended exam question.

Read the image carefully, including all question text, diagrams, charts, tables, labels, formulas, symbols, numbers, and units.

Answer in the same language as the question when possible.
Use concise wording.
Preserve exact numbers, symbols, names, and units.
If the answer is a term, number, option text, or short phrase, output only that term, number, option text, or phrase.

Think internally if needed, but output only the concise final answer.
Do not output explanations, reasoning, chain of thought, Markdown, XML tags, or <think> tags.
/no_think
\end{lstlisting}

\subsection{Visual OpenQA OCR Prompt}
\begin{lstlisting}
You are answering a multilingual visual open ended exam question.

Use the image as the primary source of truth. Read all visible question text, diagrams, charts, tables, labels, formulas, symbols, numbers, and units.

External OCR text may be provided after the prompt. Use OCR only as supporting evidence for small or blurry text. If OCR conflicts with the image, trust the image.

Answer in the same language as the question when possible.
Preserve exact numbers, units, names, formulas, and symbols.
If the answer is a term, number, option text, or short phrase, output only that term, number, option text, or phrase.
If the question asks for several parts, return the parts in a compact list in the same order.

Output only the concise final answer.
Do not output reasoning, explanation, Markdown, XML tags, citations, or <think> tags.
/no_think
\end{lstlisting}

\section{Ensemble Algorithms}
\label{app:ensemble-algorithms}

The following equations summarize our ensemble methods. Each algorithm is
described briefly in plain language, followed by its mathematical form. Let
\(\mathcal{L}=\{A,B,C,D,E\}\) be the MCQ label set.

\subsection{Algorithm 1: MCQ Weighted Score Fusion}
Each model assigns a score to every answer label. We multiply each model's
scores by its weight, sum them, and choose the largest score:
\begin{equation}
    S(\ell) = \sum_{m=1}^{M} w_m s_{m,\ell}.
\end{equation}
\begin{equation}
    \hat{\ell} =
    \mathop{\mathrm{arg\,max}}_{\ell \in \mathcal{L}} S(\ell).
\end{equation}
Here, \(w_m\) is the model weight and \(s_{m,\ell}\) is the score from model
\(m\) for label \(\ell\).

\subsection{Algorithm 2: MCQ Majority Vote with Tie Priority}
Each model votes for one label. If several labels tie, we use a fixed priority
order of models to break the tie:
\begin{equation}
    v(\ell) = \sum_{m=1}^{M} \mathbf{1}[\hat{\ell}_m = \ell].
\end{equation}
\begin{equation}
    \mathcal{T} =
    \{\ell \in \mathcal{L}: v(\ell)=\max_{\ell' \in \mathcal{L}} v(\ell')\}.
\end{equation}
\begin{equation}
    \hat{\ell} = \hat{\ell}_{\pi_k},
    \quad
    k = \min\{j: \hat{\ell}_{\pi_j} \in \mathcal{T}\}.
\end{equation}
Here, \(\mathcal{T}\) is the tied winner set and \(\pi\) is the model priority
order. If there is no tie, \(\mathcal{T}\) has one label.

\subsection{Algorithm 3: OpenQA Position Weighted Ensemble}
At each answer position, we group similar model answers, sum the weights of
models supporting each group, and choose the strongest group:
\begin{equation}
    J(a,b) =
    \frac{|T(a) \cap T(b)|}{|T(a) \cup T(b)|}.
\end{equation}
\begin{equation}
    W_p(\mathcal{C}_k) =
    \sum_{m:\,a_{m,p}\in\mathcal{C}_k} w_m.
\end{equation}
\begin{equation}
    \mathcal{C}^{*}_{p} =
    \mathop{\mathrm{arg\,max}}_{\mathcal{C}_k} W_p(\mathcal{C}_k),
\end{equation}
\begin{equation}
    \hat{a}_p = a_{m^*,p},
    \quad
    m^* =
    \mathop{\mathrm{arg\,max}}_{m:\,a_{m,p}\in\mathcal{C}^{*}_{p}} w_m.
\end{equation}
Here, \(a_{m,p}\) is the answer from model \(m\) at position \(p\), \(T(a)\) is
the token set of answer \(a\), and clusters use \(J(a,b)\geq0.35\).

\subsection{Algorithm 4: OpenQA Top 2 Union}
We keep the Qwen3-VL-32B answer by default. We replace it only when Qwen3-VL-8B
disagrees and the Qwen2.5-VL-32B verifier supports the 8B answer:
\begin{equation}
    \hat{a}_p =
    \begin{cases}
        a^{(32B)}_p,
        & J(a^{(32B)}_p,a^{(8B)}_p) \geq \tau,\\
        a^{(8B)}_p,
        & J(a^{(32B)}_p,a^{(8B)}_p) < \tau
          \ \mathrm{and}\ J(a^{(V)}_p,a^{(8B)}_p) \geq \tau,\\
        a^{(32B)}_p,
        & \mathrm{otherwise}.
    \end{cases}
\end{equation}
Here, \(\tau=0.30\), \(a^{(32B)}_p\) is the Qwen3-VL-32B answer,
\(a^{(8B)}_p\) is the Qwen3-VL-8B answer, and \(a^{(V)}_p\) is the verifier
answer.

\subsection{Algorithm 5: OpenQA Weighted Union}
We pool all unique answers, merge similar ones, rank answer clusters by total
support weight, and return the top \(N^*\) clusters:
\begin{equation}
    W(\mathcal{C}_k) =
    \sum_{m:\,\mathcal{A}_m \cap \mathcal{C}_k \neq \emptyset} w_m.
\end{equation}
\begin{equation}
    \hat{\mathcal{A}} =
    \left[
    \operatorname{rep}(\mathcal{C}_k):
    \mathcal{C}_k \in \mathrm{Top}_{N^*}\{W(\mathcal{C}_1),\ldots,W(\mathcal{C}_K)\}
    \right].
\end{equation}
Here, \(\mathcal{A}_m\) is the answer set from model \(m\), clusters are merged
using \(J(a,b)\geq0.40\), and \(N^*\) is the answer count selected by weighted
majority vote.

\bibliography{references}

\end{document}